\documentclass[11pt,a4paper]{article}
\usepackage{times,latexsym}
\usepackage{url}
\usepackage{graphicx}
\usepackage[dvipsnames]{xcolor}
\usepackage[T1]{fontenc}

\usepackage[acceptedWithA]{tacl2021v1}
\usepackage{xspace,mfirstuc,tabulary}

\newif\iftaclinstructions
\taclinstructionsfalse 
\iftaclinstructions
\renewcommand{\confidential}{}
\renewcommand{\anonsubtext}{(No author info supplied here, for consistency with
TACL-submission anonymization requirements)}
\newcommand{\instr}
\fi

\iftaclpubformat 

\else

\fi

\definecolor{km}{HTML}{FF0000}

\definecolor{ms}{HTML}{9F33FF}

\title{Reading Subtext: Evaluating Large Language Models on Short Story Summarization with Writers
\\
\definecolor{dred}{RGB}{220, 50, 32}
\definecolor{dblue}{RGB}{0, 90, 181}
{\footnotesize \textcolor{dblue}{Warning: This paper contains examples of artistic work that may include shocking or disturbing details.}}}

\author{
  Melanie Subbiah$^1$
  \and
  Sean Zhang$^1$
 \and
  Lydia B. Chilton$^1$
      \and
  Kathleen McKeown$^1$
\\
  \ \\
  $^1$Department of Computer Science, Columbia University, USA\\
  \texttt{\{m.subbiah,srz2116\}@columbia.edu}, \texttt{\{chilton,kathy\}@cs.columbia.edu}
}

\date{}

\begin{document}
\maketitle
\begin{abstract}
We evaluate recent Large Language Models (LLMs) on the challenging task of summarizing short stories, which can be 
lengthy, and include nuanced subtext or scrambled timelines. Importantly, we work directly with authors to ensure that the stories have not been shared online (and therefore are 
unseen by the models), and to obtain informed evaluations of summary quality using judgments from the authors themselves. Through quantitative and qualitative analysis grounded in narrative theory, we compare GPT-4, Claude-2.1, and LLama-2-70B. We find that all three models make faithfulness mistakes in over 50\% of summaries and struggle with specificity and interpretation of difficult subtext. 
We additionally demonstrate that LLM ratings and other automatic metrics for summary quality do not correlate well with the quality ratings from the writers.
\end{abstract}

\section{Introduction}
Narrative is a fundamental part of how we communicate and make sense of our experiences. As \citet{herman2009basic} 
describes, ``Narrative roots itself in the lived,
felt experience of human or human-like agents interacting in an ongoing way with their surrounding environment..." Understanding narrative is, therefore, a necessary skill Large Language Models (LLMs) need in order to engage with the subtleties of 
human experience and communication. 
We test how well LLMs understand the subtleties of narrative through the task of narrative summarization, with a focus on interpreting themes and meaning in addition to capturing plot. 


For our narrative form, we use fiction short stories as they present several interesting challenges. Fiction writing does not follow a clear intro-body-conclusion format and instead may follow a nonlinear timeline, hint at things only abstractly, or deliberately mislead the reader. Fiction may use multiple language varieties and creative language. For example, consider this quote from Toni Morrison's \textit{Beloved}: ``The pieces I am, she gather them and give them back to me in all the right order" \cite{morrison2004beloved}. This sentence uses complex metaphor and African American Language \cite{deas2023evaluation, grieser2022black} from the late 1800's to express a beautiful relationship between two characters. 
Finally, fiction short stories can be longer than an LLM's context window. 
Here, we consider stories up to 10,000 tokens long, which fit within the context window of the paid LLMs we use but are too long for the context window of the open-source LLM we use.

Evaluations of narrative summarization on long documents have been scarce due to several key challenges: 1) Narrative text is generally either in the public domain (and therefore likely in LLM training data) or under copyright, and 2) Holistic summary evaluation has been prohibitively difficult due to a lack of reliable automatic metrics \cite{fabbri2021summeval, chang2023booookscore} and complications with human evaluation. For example, it can take someone over an hour to thoroughly read and evaluate just one story and summary, which quickly becomes expensive. 

\citet{chang2023booookscore} make progress on these issues by purchasing recent books (which are less likely to be in models' training data but may still be discussed in the training data), and developing an LLM-based metric for evaluating summary coherence. 
We take this a step further by working directly with experienced creative writers, and thus are able to: 1) use stories that are not discussed or present in training data, 2) expand beyond evaluation of coherence to aspects of narrative understanding like faithfulness and thematic analysis, and 3) use human rather than model judgments.

We evaluate three LLMS -- GPT-4, 
Claude-2.1, and LLama-2-70B 
-- on 25 short stories. We ask authors to evaluate the summaries of their own unpublished stories\footnote{We release our code, the writer evaluation responses, and the story/summary errors here: \url{https://github.com/melaniesubbiah/reading-subtext}
} since they are experts in what they have written, focusing on coherence, faithfulness, coverage of important details, and useful interpretation in the summaries. We then present an analysis of their evaluation. 

\textbf{The key features of our work are:}\\
\textbf{1.)} Span-level, summary-level, and story-level evaluation of LLM summaries of short stories.\\
\textbf{2.)} Experienced creative writers as evaluators and short stories unseen by LLMs.\\
\textbf{3.)} Exploration of LLM ability to analyze and interpret narrative.

\textbf{Our key findings are:}\\
\textbf{1.)} GPT-4 and Claude can produce excellent summaries, but only about half the time.\\
\textbf{2.)} LLMs struggle with specificity, interpreting subtext, and unreliable narrators.\\
\textbf{3.)} LLM judgments cannot replace skilled human evaluators for this task.

Lastly, this work demonstrates the mutual benefit of working directly with communities who have valuable data as an increasing amount of online content is consumed or generated by models. 


\section{Related Work}
\textbf{Narrative Summarization.} Interest in long narrative summarization has steadily grown in the last several years. \citet{ladhak2020exploring} first introduced the task of summarizing chapters from novels, which was then expanded into the full BookSum dataset \cite{kryscinski2021booksum} and used in early studies of RLHF \cite{wu2021recursively}. Most similar to our work is SQuALITY \cite{wang2022squality}, which also focuses on short stories. However, their stories are sourced from Project Gutenberg which is likely memorized by LLMs. Across other areas of narrative understanding, recent datasets have been proposed in screenplays \cite{chen2021summscreen}, poetry \cite{mahbub2023unveiling}, 
and theory of mind evaluation \cite{xu2024opentom}.

\textbf{Summarization Evaluation.} Evaluating summary quality in any of these areas is a challenge. Work such as \citet{fabbri2021summeval} has shown the flaws in many of the traditional automatic metrics, prompting a move toward model-based reference-free methods and fine-grained span analysis. LongEval \cite{krishna2023longeval}, QAFactEval \cite{fabbri2021qafacteval}, FALTE \cite{goyal-etal-2022-falte}, and FActScore \cite{min2023factscore} have furthered methods in faithfulness, while work like SNaC \cite{goyal2022snac} and BooookScore \cite{chang2023booookscore} have focused on coherence. \citet{chang2023booookscore} is most similar to our work, but they focus on a GPT prompting strategy for evaluation, evaluate on recently published books, and only evaluate summary coherence. Several studies have benchmarked LLMs on summarization tasks \cite{tang2024tofueval, zhang2023mug, tang2022understanding, JAHAN2024108189, liu2023benchmarking, pu2023summarization}, including ones \cite{zhang2023benchmarking, goyal2023news} that have shown performance is saturated on the commonly used CNN/DM news summarization dataset \cite{hermann2015teaching}.

\label{sec:relatedwork}
\textbf{Studies with Writers.} There have been some studies that collaborate with writers to address the evaluation problem. However, these all focus on narrative generation or collaborative writing and most use amateur writers or crowdworkers rather than skilled professionals \cite{zhong2023fiction, yuan2022wordcraft, begus2023experimental, chakrabarty2022help, padmakumar2023does, yeh2024ghostwriter}. Most relevant to our work are these studies that also use professional writers but focus on narrative generation instead of summarization:\\
\citet{chakrabarty2023art} -- 10 writers, 12 stories\\ \citet{chakrabarty2023creativity} -- 17 writers, 30 stories\\ 
\citet{ippolito2022creative} -- 13 writers, 13 stories\\
\citet{huang2023inspo} -- 8 writers, 8 stories\\
These studies use similar numbers of writers and stories as us given the challenges/cost involved.

\section{Writers and Data}

To avoid using stories that may have contents or analysis available online, we find skilled writers with unpublished stories they are willing to share. We recruit writers through MFA listservs, posts on X, and direct emails. Once writers express interest, we screen for skill by 
writer education level and/or portfolio. Finally, we obtain their informed consent by sharing an infosheet on the study, which explains the task of sharing their stories and evaluating summaries, and includes how their work will be used as outlined in IRB protocol AAAU8875 (see full infosheet in Appendix \ref{sec:consent}). We ask two questions to determine the background of the writers: 1) Do you have an undergraduate or MFA writing degree?, and 2) Have you previously published any of your writing? To protect writers' anonymity, we use anonymous IDs to collect and store all the data. In line with \citet{chakrabarty2023art} and \citet{ippolito2022creative}, who use 8 and 13 writers respectively, our 
recruited group consists of 9 skilled writers -- 8/9 possess a writing degree, and 7/9 have previously published writing.

Each writer is given the opportunity to submit 
short stories that they have written and not published anywhere online (see interface in Appendix \ref{sec:interface}). We limit the writers to five stories at most, so one writer does not dominate the data. Four writers choose to submit five stories, and five submit one story, resulting in a dataset of 25 stories. We compensate the writers for their submitted stories and evaluations. Table \ref{tab:datasets} shows some summary statistics for this data. None of the stories exceed 9900 tokens in length, and we show statistics by different story length buckets: short (<3300 tokens), medium (3300-6600 tokens), and long (6600-9900 tokens). Token count is computed for each story as an average of the GPT-4 tokenizer\footnote{\url{https://cookbook.openai.com/examples/how_to_count_tokens_with_tiktoken}} count and the Llama2 tokenizer\footnote{\url{https://huggingface.co/docs/transformers/model_doc/llama2\#transformers.LlamaTokenizer}} count. Claude does not have a publicly available tokenizer, so we cannot include its token count in this average.

We also explore how similar the stories may be to the LLM training data by measuring their perrplexity. We cannot compute perplexity using the GPT-4 and Claude APIs, so we get an approximate number by using Llama-2-7B (Float16), which we can run on our own compute. This model assigns perplexity scores between 9.2 and 21.6 to the stories. For reference, the first chapter of \textit{Pride and Prejudice} (a story in Llama-2's training data) has a perplexity score of 1.3, which is just under the model's training loss. These scores therefore validate that the stories are unfamiliar to the models.

\section{Summary Generation}
\begin{table}[t]
\centering
\footnotesize
\begin{tabular}{l|cc|ccc}
\hline
\textbf{Length} & \multicolumn{2}{c|}{ \textbf{Stories}} & \multicolumn{3}{c}{ \textbf{Summary Avg. Len.}}\\
 \textbf{Bucket} & \# &  Avg. Len. &  GPT4 &   Claude & \small Llama \\
\hline
 \scriptsize Short & 10 & 1854 & 487 & 397 & 458\\
\scriptsize Medium & 9 & 4543 & 500 & 339 & 482\\
\scriptsize Long & 6 & 8126 & 531 & 382 & 592\\\hline
 \textbf{Total} & 25 & 4327 & 502 & 373 & 499\\\hline
\end{tabular}
\caption{For each story length bucket, we show the count (number of stories) and average length (token count) of the stories, and the average length (word count) of the summaries generated by each model for those stories.}
\label{tab:datasets}
\end{table}

\begin{figure}[t]
    \centering
    \includegraphics[width=\columnwidth]{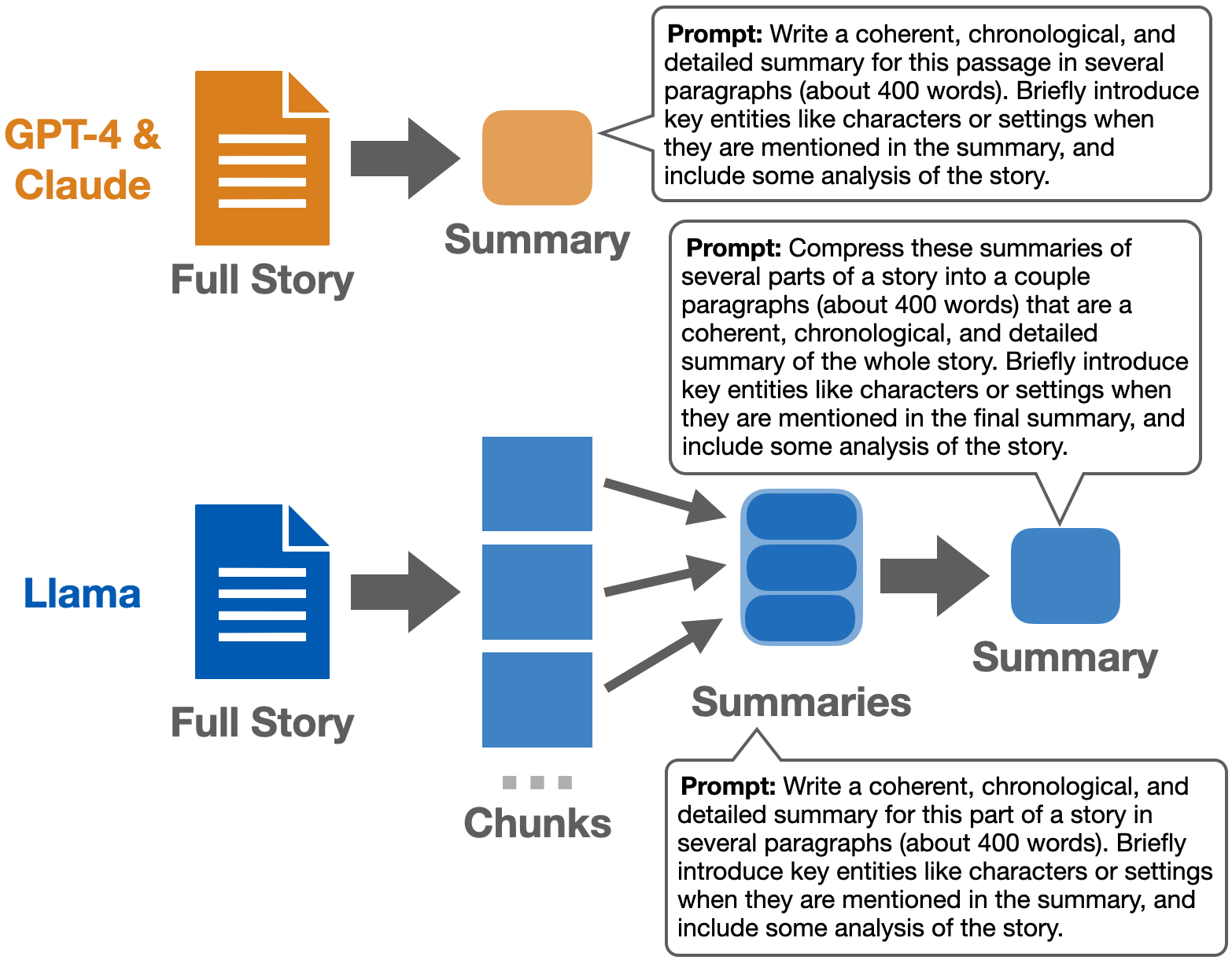}
    \caption{The two different methods we use for summarization and the associated prompts for the models. GPT-4 and Claude have sufficient input context to summarize a whole story, whereas Llama has to use a chunk-then-summarize approach for longer stories.}
    \label{fig:methods}
\end{figure}

\begin{table*}[t]
\centering
\small
\begin{tabular}{l|ccc|cccc|c}
\hline
 \textbf{Model} & \textbf{Coverage} & \textbf{Density} & \textbf{Compression} & \textbf{2-grams} & \textbf{3-grams} & \textbf{4-grams} & \textbf{5-grams} & \textbf{LCS} \\
\hline
GPT-4 & 72.52\% & 1.33 & 7.45 & 22.53\% & 5.18\% & 1.76\% & 0.71\% & 5.72\\
Claude & 73.65\% & 1.26 & 10.86 & 19.80\% & 3.96\% & 1.25\% & 0.50\% & 4.16\\
Llama & 66.09\% & 1.08 & 7.66 & 16.33\% & 3.04\% & 0.96\% & 0.32\% & 4.40\\\hline
\end{tabular}
\caption{We report a variety of metrics for the summaries in relation to the stories: coverage, density, compression, percent of n-gram overlap, and word count of the longest common substring. Each metric is averaged across the summaries generated by each model.}
\label{tab:datametrics}
\end{table*}

We evaluate automatic summarization using three recent LLMs: \textbf{GPT-4} (November 2023 update), \textbf{Claude-2.1}, and \textbf{Llama-2-70B-chat} (instruction-tuned for chat). GPT-4 \cite{openai2023gpt4} and Claude-2.1\footnote{\url{https://www.anthropic.com/news/claude-2-1}} have very long input contexts and can ingest an entire short story. Llama-2 \cite{touvron2023llama} serves as a comparison point for smaller open-source models. \citet{chang2023booookscore} also used these three LLMs in their work on automatic coherence evaluation in long narrative summarization.

Protecting the writers' unpublished work is paramount when using each of these three models. We do not want their stories to be saved or trained on by OpenAI, Anthropic, or HuggingFace. We inform the writers of this risk in the consent form (see Appendix \ref{sec:consent}) and complete the available forms/settings with each of these companies to request our data not be stored for long periods or used for training. 

For each story, we then generate three summaries -- one from each of the three models. For GPT-4 and Claude, we use one prompt (see Figure \ref{fig:methods}) as they both have a context length long enough to process a whole short story. Claude refuses to summarize two of the stories which do not meet its content restrictions\footnote{For all of the Claude results, we remove the values for these two summaries to show representative numbers for the summaries it did produce. We also remove one coherence score which was inaccurate due to a bug in the interface.}. One of these stories is about a shooting and the other involves sex and robbery.

Given Llama's short context window, we use hierarchical merging of chunk-level summaries \cite{chang2023booookscore} for stories in the medium and long length buckets. Depending on length, we chunk each story into 1-4 chunks using section or paragraph breaks. Llama summarizes each of these chunks separately and then summarizes the concatenation of these summaries (see Figure \ref{fig:methods}). 

For all three models, we access them from December 2023 through January 2024, and we use simplified versions of \citet{chang2023booookscore}'s prompts (see Figure \ref{fig:methods}). For GPT-4\footnote{\texttt{gpt-4-1106-preview} on \url{https://platform.openai.com/}} and Claude\footnote{\texttt{claude-2.1} on \url{https://console.anthropic.com}}, we use \texttt{max\_tokens}=1000 and \texttt{temperature}=0.3. For Llama\footnote{\texttt{meta-llama/Llama-2-70b-chat-hf} on \url{https://huggingface.co/}}, we use the default settings in HuggingChat\footnote{\url{https://huggingface.co/chat/}}.

Table \ref{tab:datasets} displays summary statistics for the generated summaries. Notice that our prompt asks models to use about 400 words and Claude undershoots but comes closest to this target while GPT-4 and Llama are 100 words over on average. See Appendix \ref{sec:prompts} for full prompting details and costs. 

In Table \ref{tab:datametrics}, we report the average coverage, density, and compression metrics for each model's summaries. Coverage is the percent of words in the summary that come from the story, density is the average length of segments that directly match the story text, and compression is how many times shorter the summary is in relation to the story \cite{fabbri2021summeval, grusky2018newsroom}. These metrics show that Llama is the least extractive and Claude compresses the information the most.

In addition to these commonly used metrics, we further investigate how much the models copy directly from the stories in Table \ref{tab:datametrics}. We report the average percent of n-grams in the summaries that match n-grams in the stories, and the average word count of the longest substring that exactly matches wording from the story. For the n-gram matching, we remove punctuation and lower-case and stem words. We see that GPT-4 copies a significant amount of wording from the stories. On average, it produces summaries with almost 6-word long exact-match substrings and it has the highest percentages of n-gram overlap. These numbers indicate that models are not copying long quotes, but they are copying a non-trivial amount of unique phrasing. None of the summaries use quotation marks to appropriately attribute copied text.

\section{Evaluation}

We compare models using a combination of span-level, summary-level, and story-level evaluation. We assess summary quality in terms of four attributes:\\
\textbf{Coverage} -- Does the summary cover the important plot points of the story?\\
\textbf{Faithfulness} -- Does the summary misrepresent details from the story or make things up?\\
\textbf{Coherence} -- Is the summary coherent, fluent, and readable?\\
\textbf{Analysis} -- Does the summary provide any correct analysis of some of the main takeaways or themes from the story?

\textit{Coverage}, \textit{faithfulness}, and \textit{coherence} are commonly evaluated in summarization (e.g., \cite{zhang2023benchmarking}), and we add \textit{analysis}, which is important for capturing narrative. Each writer is shown the summaries of their own stories to evaluate as they are deeply familiar with the contents and can therefore judge aspects like \textit{faithfulness} and \textit{analysis} quickly and accurately. 

\subsection{Span-Level Error Categorization}
We ask the writers to highlight spans in the summaries they view as errors and categorize them (see Appendix \ref{sec:doccano} for interface and cost). We do not make this task mandatory as it requires significantly more time from the writers than the summary-level evaluation discussed next. 7/9 writers choose to participate, which results in 69/75 summaries with span-level annotations.

For faithfulness, we use error categories inspired by elements from narrative theory for evaluating narrative understanding in children \cite{xu2022fantastic, paris2003assessing, mandler1977remembrance}. 
We determine these categories are well-defined as in prior work by asking three NLP researchers to categorize a sample of 60 faithfulness errors. 
These annotators achieve moderate inter-annotator agreement for the labels (Fleiss-Kappa of .51), indicating the categories are valid. For coherence, we use categories defined and validated by \citet{chang2023booookscore}. For coverage, we cannot highlight spans that should have been included, so we can only focus on things that should not have been included or are non-specific. The full list of error categories are defined as:\\
\textbf{Coverage}

\textsc{Insignificant} - Does not need to be included and makes the summary less readable

\textsc{Vague} - Important but covered in a vague way\\
\textbf{Faithfulness}

\textsc{Feeling} - Inaccurate about a character's emotions/reaction/internal state, incorrectly answers a question like "How did X react" or "What was X thinking?"

\textsc{Character} - Inaccurate about the identity or nature of a character, incorrectly answers a question like "How would you describe X?" or "Who is X?"

\textsc{Causation} - Inaccurate about the causal relationship of events, incorrectly answers a question like "Why did Y happen?" or "What was the result of Y?"

\textsc{Action} - Inaccurate about the behavior of a character, incorrectly answers a question like "What did X do?"

\textsc{Setting} - Inaccurate about the details of the story world and the time/place of events, incorrectly answers a question like "Where/when did X happen?" or "What is the setting of the story?"\\
\textbf{Coherence}

\textsc{Inconsistent} - Inconsistent with other details in the summary

\textsc{Abrupt Transition} - Transitions suddenly to a new scene without a relevant connection

\textsc{Missing Context} - Introduces a new character, event, or object without enough context/detail to understand it

\textsc{Repetition} - Unnecessary repetition of detail\\
\textbf{Analysis}

\textsc{Unsupported} - Interprets the story but the conclusions do not make sense with or are unsupported by the story

\subsection{Summary-Level Ratings}
\begin{figure*}[t]
    \centering
    \includegraphics[width=0.95\textwidth]{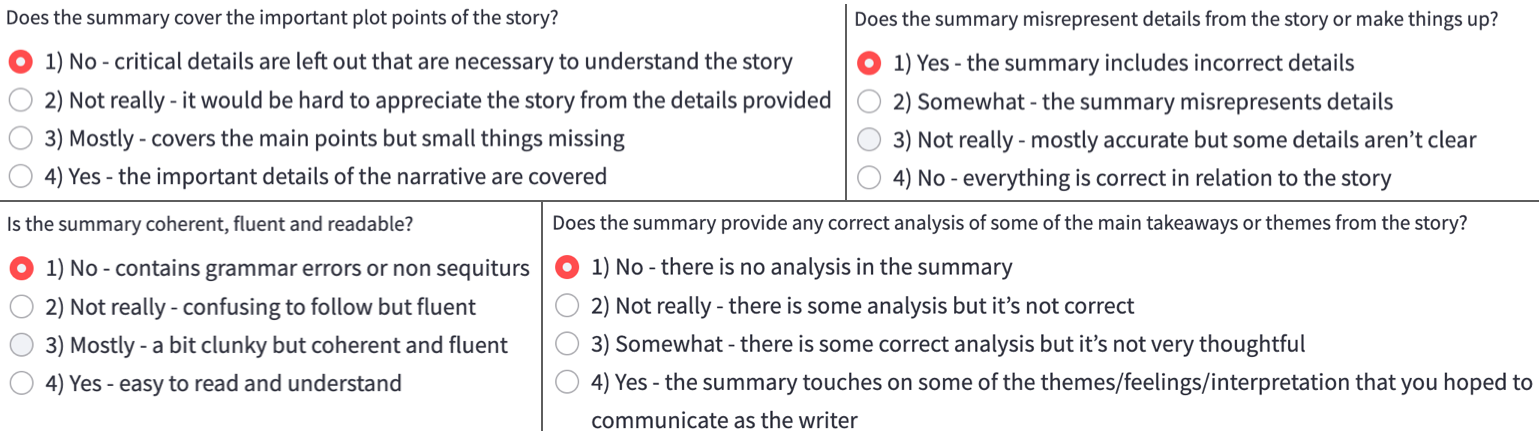}
    \caption{Interface screenshots showing the questions writers are asked to evaluate the summaries of their stories using a 4-point Likert scale.}
    \label{fig:questions}
\end{figure*}

\begin{table}[t]
\centering
\footnotesize
\begin{tabular}{c|cccc}
\hline
\scriptsize Rating & Cover. & Faithful. & Coher. & Analys.\\\hline
1 & 1.00 & 3.33 & - & 0.00\\
2 & 1.92 & 1.32 & 1.00 & 2.33\\
3 & 1.48 & 0.87 & 1.36 & 1.50\\
4 & 0.50 & 0.89 & 0.76 & 1.22\\\hline
\end{tabular}
\caption{Average number of span-level errors by attribute labeled in summaries given each Likert score rating for that attribute.}
\label{tab:ratingerror}
\end{table}

 We ask the writers to rate each of the four attributes on a Likert scale from 1 to 4 with some guidance on what each score means (see Figure \ref{fig:questions}, full interface and cost shown in Appendix \ref{sec:interface}). This step is completed before the span-level annotation to keep conditions equal between writers who opt in or out of the span-level annotation.

 To help interpret the Likert scores, in Table \ref{tab:ratingerror}, we present a breakdown of the average number of span-level errors annotated in each attribute when a summary is given a certain Likert score rating for that attribute. We can see that for most attributes, there is an increase in average error count for an attribute as the rating assigned to that attribute decreases. The number of 1-ratings is quite small for many attributes so the relationship between number of errors and rating is noisier for that bucket. Overall, these numbers support our 4-point Likert scale and demonstrate that the summary-level ratings correspond to meaningful differences in the number of errors in a summary.

Once each writer has read and evaluated each summary for a story, we ask them to rank the three summaries from the three different models in order of their preference. In addition to the ratings and ranking, the writers provide open-ended feedback on each summary, which we include quotes from throughout our discussion of the results.

We also explore automatic metrics for rating summaries. We try simple automatic metrics, using ROUGE \cite{lin2004rouge} and BERTScore \cite{zhang2019bertscore} against the full story as a reference. We try recent attribute-specific metrics designed for faithfulness: AlignScore \cite{zha2023alignscore}, UniEval \cite{zhong2022towards}, and MiniCheck \cite{tang2024minicheck}. Finally, we ask GPT-4 and Claude (Llama's context window is too short) the four questions shown in Figure \ref{fig:questions} with the same wording to see how their answers compare to the opinions of the writers (see Appendix \ref{sec:prompts} for prompts). We run BooookScore and FABLES as another comparison point for LLM-based coherence and faithfulness evaluation respectively \cite{chang2023booookscore, kim2024fables}. For FABLES, we average the labels for all the claims in a summary to get a score for the whole summary. 

\subsection{Story-Level Style Effects}

\begin{figure}[t]
\centering
\includegraphics[width=\columnwidth]{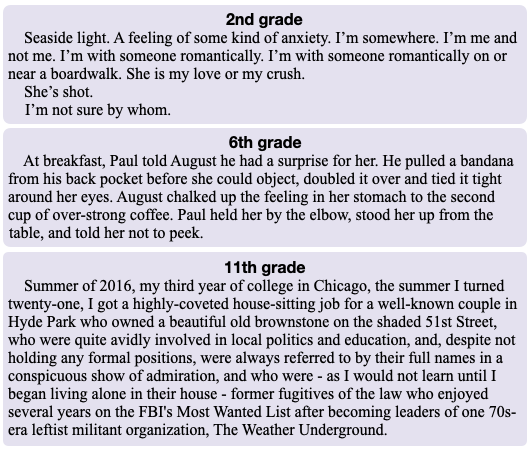}
\caption{Examples of openings from stories scored at different reading-levels by the Flesch-Kincaid score.}
\label{fig:reading}
\end{figure}

We examine how writing style at the story level affects model summarization to see if certain types of stories are harder for models to summarize well. First, we use Genette's model of narrative elements \cite{piper2021narrative, genette1980narrative}:\\
\textbf{Narrating} -- Narrator's influence on style\\
\textbf{Story} -- All the events implied by the narrative\\
\textbf{Discourse} -- Ordering/inclusion of explicit events

\label{sec:unreliable}
At the \textit{narrating} level, we compare stories with an unreliable vs. reliable narrator. An unreliable narrator communicates something different from what the reader is meant to perceive \cite{booth1983rhetoric}. For example, one narrator states early in the story, "\textit{If I were to describe myself, I would say that I am practical. I try to be logical, but that isn't exactly what I mean. Here's another way to put it: there are two kinds of people in the world, me and my brother Jack.}" The narrator says many things without really saying anything, which gives us a clue that he may not actually be logical and practical. At the \textit{story} level, we compare stories with a detailed subplot involving niche knowledge against those which focus on commonplace settings. For example, one detailed story centers on the main character sleeping with her sister's boyfriend, but the subplot to this emotional arc involves a lot of ancient greek history. At the \textit{discourse} level, we compare stories with flashbacks against those with a linear timeline. We hypothesize that reliable narrators, fewer details, and linear plots will be easier to summarize. The writer of each story verifies the labels for the story in these three categories.


Finally, we analyze the complexity of the story wording using the Flesch-Kincaid readability test \cite{kincaid1975derivation}. This score estimates the number of years of formal education (grades 1 through 12) one might need to understand the writing easily. It is based on average word and sentence length, so while it captures something about the complexity of the wording, it is in no way a measure of the overall quality of the writing. We explore whether the reading-level of the writing affects the summary quality (see examples of writing with assigned grade levels in Figure \ref{fig:reading}).

\section{Results}


\subsection{How good are the summaries?}

\begin{table}[t]
\centering
\footnotesize
\begin{tabular}{c|cccc|c}
\hline
Model & \scriptsize Cover. & \scriptsize Faithful. & \scriptsize Coheren. & \scriptsize Analys. & Avg.\\
\hline
\scriptsize GPT-4 & 3.48 & 3.12 & 3.52 & 3.40 & \textcolor{OliveGreen}{\textbf{3.38}} \\
\scriptsize Claude & 3.17 & 2.65 & 3.41 & 3.26 & \textcolor{OliveGreen}{\textbf{3.12}} \\
\scriptsize Llama & 2.40 & 1.92 & 3.08 & 2.76 & 2.54\\\hline
\scriptsize GPT-4 & 56\% & 44\% & 60\% & 56\% & 54\%\\
\scriptsize Claude & 39\% & 30\% & 59\% & 43\% & 43\%\\
\scriptsize Llama & 12\% & 8\% & 32\% & 20\% & 18\%\\\hline
\end{tabular}
\caption{Average scores assigned to each attribute of a summary by the writers. The first set of rows are the averaged raw scores out of 4, and the second set of rows are the percent of summaries given a perfect score of 4.}
\label{tab:scores}
\end{table}

\begin{figure*}[t]
  \centering
  \includegraphics[width=\textwidth]{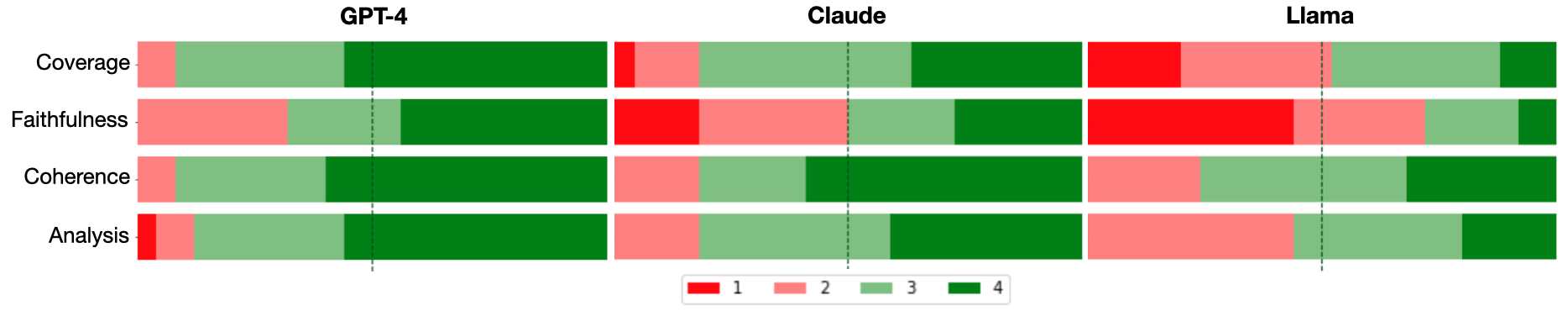}
  \caption{Distribution of Likert score ratings for each model's summaries by attribute.}
  \label{fig:horizontal}
\end{figure*}

\begin{figure}[t]
  \centering
  \includegraphics[width=\columnwidth]{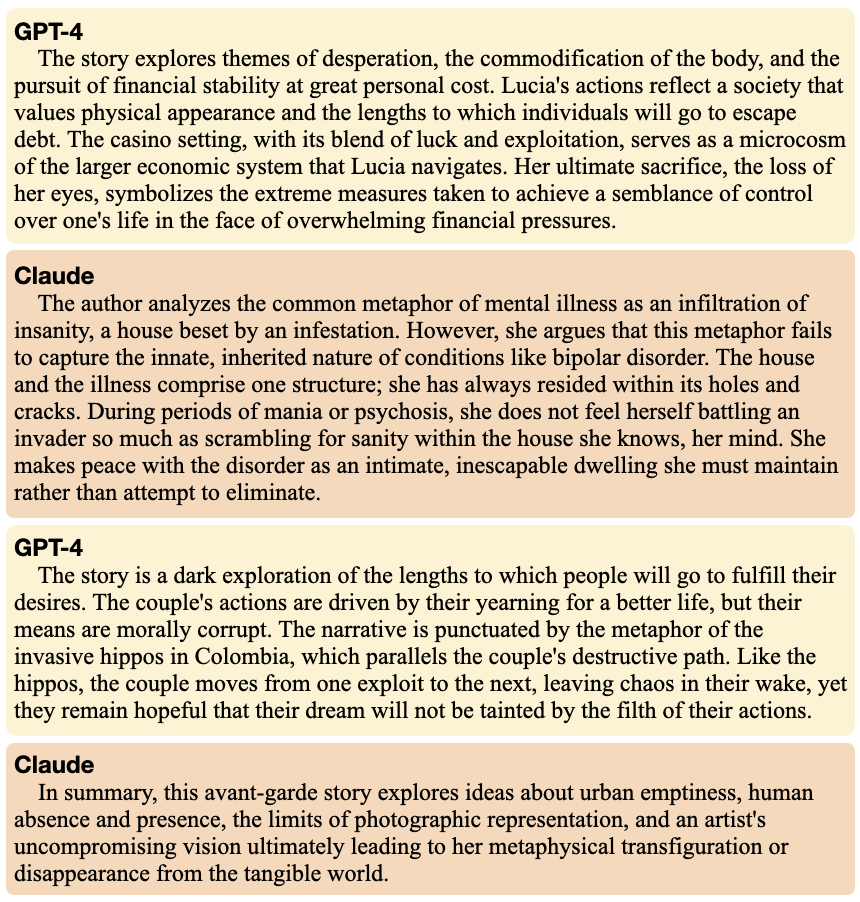}
  \caption{Examples of some of the best analysis-focused sentences from GPT-4 and Claude summaries.}
  \label{fig:analysis}
\end{figure}

\begin{figure}[t]
  \centering
  \includegraphics[width=.9\columnwidth]{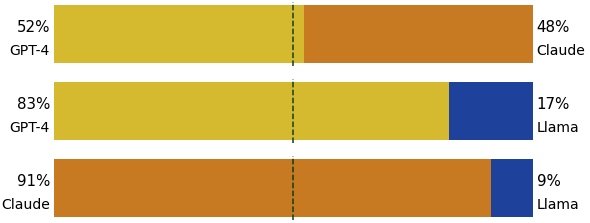}
  \caption{Percent of each model's summaries that are ranked higher than another's in the 3-way ranking the writers are asked to do. Yellow indicates GPT-4, orange is Claude, and blue is Llama.}
  \label{fig:rankings}
\end{figure}

In Table \ref{tab:scores}, we can see that the models are capable of producing good summaries. Many of the average scores are >3, and there is a percentage of summaries in each attribute (coverage, faithfulness, coherence, and analysis) that receive a perfect rating of 4. GPT-4 summaries have the highest scores across all four attributes. However, by percentage of perfect scores, GPT-4 still makes mistakes on 46\% of the summaries on average across attributes. Faithfulness is the lowest score for all three models, with only 8\% rated fully correct for Llama, 30\% for Claude, and 44\% for GPT-4. Llama performs significantly worse than the other two models on all attributes (Wilcoxon signed-rank test, $p<.05$). A further breakdown of rating distribution is shown in Figure \ref{fig:horizontal}.

Interestingly, we find that while models have the appearance of fluency, they are given a less than perfect score for coherence 40\% or more of the time as there is more to a coherent narrative summary than just fluency. For analysis, writers felt that 56\% of the GPT-4 summaries summarized some of the themes and interpretation they had hoped to communicate as a writer, which is a challenging task even for humans (see examples of the best analysis in Figure \ref{fig:analysis}). However, in Table \ref{tab:ratingerror}, we see that summaries rated 4 for analysis actually contain at least one error in analysis on average. Furthermore, by breakdown of span-level error counts in Table \ref{tab:allerrors}, we see that writers find as many errors in analysis as in faithfulness. These numbers suggest that while writers may have been impressed by the models' ability to do any analysis in their ratings, analysis remains a challenge.

In Figure \ref{fig:rankings}, we present a pairwise comparison of what percentage of the time one model outranked another when the writers ranked the summaries as first, second, and third. Claude and GPT-4 are preferred over each other equally. Claude is almost always preferred over Llama, and Llama is rarely ranked higher than the other two models. This suggests the writers evaluate both GPT-4 and Claude relatively equally (GPT-4 has higher scores, but Claude is ranked higher on average), which is also supported by GPT-4's scores not being significantly better than Claude's (Wilcoxon signed-rank test, $p>.05$). It is also important to remember though that Claude refused to summarize two of the stories.

Overall, \textbf{GPT-4 and Claude can produce excellent summaries but only about half the time}. On the best summaries, the writers commented things like: "\textit{it did analysis that even I — the writer — hadn't done! Very clever.}", "\textit{I'm stunned at the thoughtfulness and thoroughness}", and "\textit{something that I would find on a Sparknotes website}".

On the worst summaries, they commented things like: "\textit{completely misses the fundamental thread}", "\textit{tries to make too many analyses like a high school english class while missing the overall bigger feelings}", and "\textit{makes use of...stock phrases or stand-ins rather than accurate, specific summary}". Also, for one of the stories Claude refused to summarize, the writer commented, "\textit{Art is such an important tool for processing. I don't like that the summary wasn't able to process a recurring vague nightmare.}"

\subsection{What mistakes do the models make?}


\begin{table}[t]
\centering
\small
\begin{tabular}{p{2.3cm}|ccc|c}
\hline
Error Type& GPT4&Claude & Llama & \textbf{Tot.}\\\hline\hline
\textbf{Coverage:}&&&&\\
insignificant & 4 & 2 & 8 & 14\\
vague & 15 & 19 & 30 & \textcolor{Maroon}{\textbf{64}}\\\hline
Sub-Total & 19 & 21 & 38 & 78\\\hline\hline
\textbf{Analysis:}&&&&\\
unsupported & 18 & 16 & 70 & \textcolor{Maroon}{\textbf{104}}\\\hline\hline
\textbf{Coherence:}&&&&\\
repetition & 3 & 4 & 0 & 7\\
missing context & 11 & 11 & 15 & \textcolor{Maroon}
{\textbf{37}}\\
inconsistent & 1 & 3 & 1 & 5\\
abrupt transition & 3 & 6 & 8 & 17\\\hline
Sub-Total & 18 & 24 & 24 & 66\\\hline\hline
\textbf{Faithfulness:}&&&&\\
feeling & 5 & 8 & 10 & 23 \\
Causation & 2 & 9 & 4 & 15\\
Action & 5 & 6 & 28 & \textcolor{Maroon}{\textbf{39}}\\
Character & 2 & 1 & 15 & 18\\
Setting & 2 & 2 & 5 & 9 \\\hline
Sub-Total & 16 & 26 & 62 & 104\\\hline\hline
\footnotesize \textbf{Total} & 71 & 87 & 194 &  \\\hline
\end{tabular}
\caption{Number of errors categorized under each narrative element by the writers for each model.}
\label{tab:allerrors}
\end{table}


\begin{figure*}[t]
  \centering
  \includegraphics[width=\textwidth]{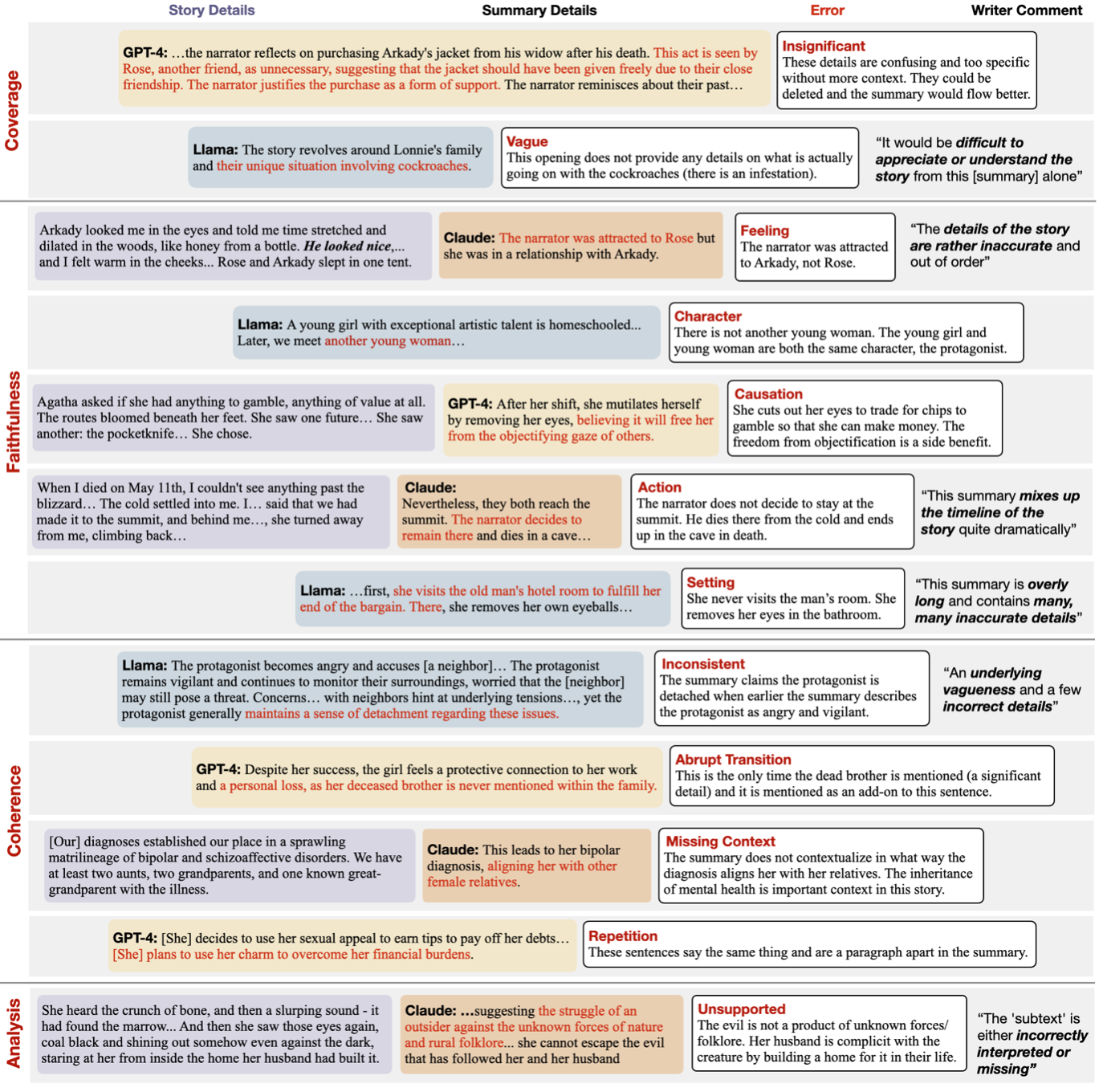}
  \caption{Examples of span-level errors assigned by the writers in each of the categories.}
\label{fig:allexamp}
\end{figure*}

In Figure \ref{fig:allexamp}, we show examples of issues models have in each attribute. In general, we find that while writers rank GPT-4 and Claude summaries somewhat equally, Claude summaries often leave out important details that lower their coverage score and create issues with faithfulness as identified by the numerical scores. For example, while the GPT-4 coverage error in Figure \ref{fig:allexamp} includes too many details, for one story, Claude leaves out an entire thematically important subplot. The narrator desperately needs money, and Claude fails to mention she is in debt due to a medical crisis.

Looking at the categorization of errors in Table \ref{tab:allerrors}, we find that \textbf{vagueness} is a significant issue across all three models. We see there is a large amount of \textbf{unsupported analysis} for all three models, and this is one of the biggest error categories overall. In coherence, the biggest error category is \textbf{\textit{missing context}}. This overlaps with coverage and lets us know that significant details are left out or too vague to fully understand. Lastly, we find that \textbf{most faithfulness errors across the models are in \textit{Action} and \textit{Feeling}}. These errors require interpretation of what characters did and how they reacted or felt. 

For example, in Figure \ref{fig:allexamp}, the \textit{Feeling} error is a misinterpretation of who the narrator is attracted to. Claude fails to interpret phrases like "\textit{He looked nice}", and "\textit{I felt warm in the cheeks}", as signs of attraction to Arkady. Another \textit{feeling} error describes a character as having a difficult relationship with their mother, when actually they are quite close, but the circumstances around their relationship have been difficult because the mother almost died from cancer. One writer comments on another summary, "\textit{The ending of contentment is supposed to be contemplation. [The summary] leaves out any feeling of hesitancy or reflection}". The \textit{action} error example in Figure \ref{fig:allexamp} mistakenly states that the narrator decided to stay at the summit. However, "\textit{The cold settled into me}", is meant to imply that the narrator died at the summit, which was not an action or choice.

Taken in conjuction, all of these results show that \textbf{models struggle with identifying the right level of specificity and with interpreting subtext}. Three writers commented on subtext specifically, saying, "\textit{Subtext is missing entirely}" on a Llama summary, "\textit{Everything that's `text' is accurately represented and well summarized, but the `subtext' is either incorrectly interpreted or missing}" on a Claude summary, and "\textit{The summary struggles with subtext}" on a GPT-4 summary. 

However, the error distribution for each model varies. For example, Llama has a high percentage of \textit{Character} errors. This is likely an artifact of the chunk-then-summarize strategy as sometimes characters are conflated or split in two as the model fails to track their details across different chunk summaries (see \textit{Character} example in Figure \ref{fig:allexamp}). \citet{chang2023booookscore} also found a similar type of error (\textit{entity omission}) to be most common with chunk-then-summarize approaches. 

It is interesting to note that some faithfulness errors come from the models being overly normative as evidenced by the Claude \textit{Feeling} error in Figure \ref{fig:allexamp}. In this case, Claude assumes a heteronormative monogamous interpretation of the interaction instead of what is written. One writer observed this phenomenon in another summary, commenting that the summary had "\textit{a few extrapolations... which presume things about the narrative based on a hypernormative interpretive framework}".

Lastly, some of the writers commented that the models would copy
unique wording from the stories without placing it in quotations, which bordered on plagiarism. We flag this for future work and include statistics on it in Table \ref{tab:datametrics}.


\subsection{Do some aspects of writing style affect summary quality?}
\subsubsection{Length}

\begin{figure*}[t]
    \centering
    \includegraphics[width=\textwidth]{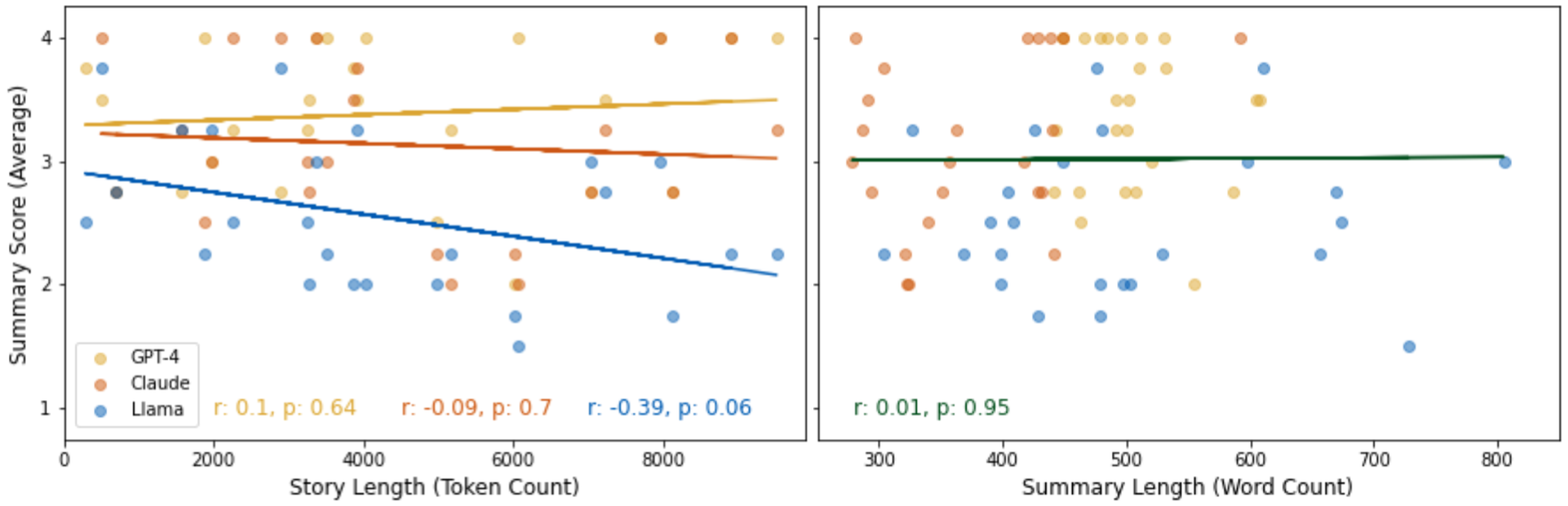}
    \caption{We plot story length (in tokens) and summary length (in word count) against writer summary ratings averaged across the four attributes. In the left plot, we show the line of best fit and Pearson's $r$ with $p$-value for each model individually. In the right plot, we show the correlation across the full set of summaries. } 
    \label{fig:lengths}
\end{figure*}


In Figure \ref{fig:lengths}, we look at the correlation between story length and summary rating (left plot). We do see a downward trend for Llama. Llama's summaries get worse as stories get longer, which is expected given the chunk-then-summarize method it employs for longer stories. However, the correlation is not quite statistically significant. For Claude and GPT-4, it seems the \textbf{long-context models can summarize short and long stories equally well} (up to 10,000 tokens). 

In Figure \ref{fig:lengths}, we also compare the average scores across different summary lengths (right plot), to check if there is a bias in ratings to shorter or longer summaries. We do not find any correlation between summary length and quality. 

\subsubsection{Reading-Level}



We plot the reading-level (estimated years of education needed to understand the wording easily) against the writer-assigned scores averaged across the four attributes for each summary. We find \textbf{no significant correlation between story reading-level and writer-assigned summary scores} for any of the models (GPT-4: $r$ .11, $p$ .61; Claude: $r$ .33, $p$ .12, Llama: $r$ .23, $p$ .27). The average Flesch-Kincaid grade-level for the group of stories is 6th grade. As can be seen in the 2nd grade example in Figure \ref{fig:reading} though, which features someone being shot, this score captures an aspect of the wording but does not consider age-appropriate content or conceptual complexity. 
\subsubsection{Narrating, Story, and Discourse}

\begin{table}[t]
\centering
\footnotesize
\begin{tabular}{r|c|ccc}
\hline
Style Element& \#&GPT4 & Claude & Llama\\
\hline
\textbf{Narrating}: reliable & 15 & \textcolor{OliveGreen}{\textbf{3.45}} & \textcolor{OliveGreen}{\textbf{3.29}} & \textcolor{OliveGreen}{\textbf{2.60}}\\
 unreliable & 10 & 3.28 & 2.89 & 2.45\\
\hline
\textbf{Story}: detail$-$ & 16 & \textcolor{OliveGreen}{\textbf{3.42}} & 2.98 & \textcolor{OliveGreen}{\textbf{2.59}}\\
detail$+$ &  9 & 3.31 & \textcolor{OliveGreen}{\textbf{3.36}} & 2.44 \\
\hline
\textbf{Discourse}: linear & 18   & 3.29 & 3.05 & \textcolor{OliveGreen}{\textbf{2.56}} \\
 nonlinear & 7 & \textcolor{OliveGreen}{\textbf{3.61}} & \textcolor{OliveGreen}{\textbf{3.32}} & 2.50  \\
\hline
\end{tabular}
\caption{Average scores for summaries of stories with different style elements. Each score is averaged across the four attributes for a summary.}
\label{tab:stypes}
\end{table}

In Table \ref{tab:stypes}, we see that stories with unreliable narrators are harder for all three models to summarize. All three models have lower average scores in this bucket. As an example of an error due to an unreliable narrator, the same unreliable narrator who is quoted in Section \ref{sec:unreliable}, further describes himself as wanting a "\textit{small, normal life}", and then describes his new love interest: "\textit{I met a girl... I think she's just like me. She loves traveling... It's pretty cool.}" GPT-4 mistakenly characterizes this interaction as: "\textit{The protagonist meets [a girl]... with whom he shares... a desire for a simple life.}" even though there is no evidence she likes a simple life (or even that we should believe the narrator that he does), other than that the protagonist has described her as "\textit{like [him]}". 

For story and discourse, the models have mixed performance across the different buckets. Llama does have lower scores in the buckets we hypothesized might be more challenging. Overall, it seems that \textbf{unreliable narrators are a challenge for LLMs}, whereas level of detail and flashbacks depend on the model. One writer commented directly on the issue of an unreliable narrator by observing, "\textit{there's a second layer here that the summary misses somewhat—the narrator is not really to be trusted}".

\subsection{Can an LLM replace the writers in evaluating the summaries?}

\begin{figure}[t]
    \centering
    \includegraphics[width=\columnwidth]{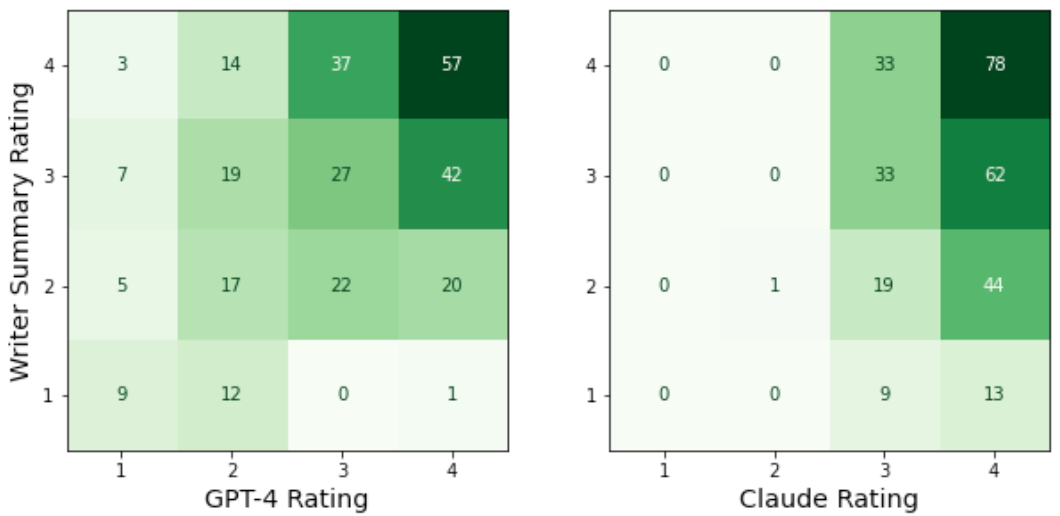}
    \caption{Confusion between model-assigned scores and human-assigned scores for GPT-4 and Claude.}
    \label{fig:confusion}
\end{figure}



\begin{table*}[t]
\centering \footnotesize
\begin{tabular}{l|cccc|cccc|ccc|c|ccc}
&\multicolumn{4}{c|}{\textbf{LLM}}&\multicolumn{4}{c|}{\textbf{ROUGE}}&\multicolumn{3}{c|}{\textbf{BERTScore}}&\textbf{Lla.2} &\multicolumn{3}{c}{\textbf{Faith. Metrics}}\\
& \scriptsize GPT4 & \scriptsize Clau.& \scriptsize BkS & \scriptsize Fab. & \scriptsize R1&\scriptsize R2&\scriptsize  RL&\scriptsize  RLS & \scriptsize Prec. & \scriptsize Rec. & \scriptsize F1 &\scriptsize  PPL &\scriptsize  AlS &\scriptsize  UnE &\scriptsize  MiC\\\hline\hline
\scriptsize Cov. & \textcolor{OliveGreen}{$.46^*$} & .18  & - & - & $.33^*$ & $.36^*$ & $.32^*$ & $.30^*$ & $.29^*$ & $.21$ & $.25^*$ & .15 & -& -  &- \\
\scriptsize Fai. & \textcolor{OliveGreen}{$.37^*$} & .05 & - & $.29^*$ & $.18$ & $.28^*$ & $.18$ & $.17$ & $.27^*$ & $.25^*$ & $.27^*$ & .00 & -.04 & $.32^*$ & $.20^*$\\
\scriptsize Coh. & .18 & .02 & .11 & - & $.18$ & $.21$ & $.16$ & $.16$ & \textcolor{OliveGreen}{$.29^*$} & -$.01$ & $.09$ & -.09 & -& - & -\\
\scriptsize Ana. & \textcolor{OliveGreen}{$.21^*$} & .02 & - & - & -$.05$ & -$.01$ & -$.06$ & -$.08$ & $.15$ & $.14$ & $.16$ & .12& -& -&-\\\hline
\scriptsize \textbf{Avg.} & - & - & - & - & $.21$ & $.28^*$ & $.20$ & $.18$ & \textcolor{OliveGreen}{$.32^*$} & $.20$ & $.26^*$ & .14& -&- &-\\
\end{tabular}
\caption{Correlation between automatic metrics and writer-assigned scores. We report Pearson's $r$ (* indicates p-value <.05). The `Avg.' row takes the average score across the four attributes as the writer score. We do not report correlation for attributes the metrics are not designed for. Metric Key for uncommon abbreviations: BkS - BooookScore, Fab. - FABLES, RLS - ROUGE-LSum, AlS - AlignScore, UnE - UniEval, MiC - MiniCheck.}
\label{tab:automatic}
\end{table*}

In Figure \ref{fig:confusion}, we compare the GPT-4 and Claude scores to the writer scores from Table \ref{tab:scores}, and we see considerable confusion across the ratings. Claude mostly only rates summaries as 3 or 4, causing overestimation of ratings (3.67 average score vs. 3.01 from the writers). 

In terms of attributes, both models rate the coherence of the summaries much higher than the writers do (average scores: GPT-4 - 3.84, Claude - 3.82, Writers - 3.33). GPT-4 largely underestimates the coverage score for all three models (average score 2.68 vs. 3.01 from the writers). Many of the writers commented that the GPT-4 summaries were too detailed, so the model may struggle with identifying what is most important to cover. 
Claude overestimates performance on faithfulness and analysis. It gives 85\% of summaries a 4 on faithfulness on average and 85\% on analysis, relative to averages of 27\% and 39\% respectively from the writers. GPT-4 also overestimates faithfulness performance just for itself (64\% scores of 4 vs. 44\% from the writers). 

We show the Pearson's $r$ correlation between the model scores and the writers' scores in Table \ref{tab:automatic} across attributes. We see no correlation for Claude scores and some statistically significant but weak correlation for GPT-4 scores (indicating there may be a relationship but the scores are not well calibrated). \textit{Coverage} is the one attribute with moderate positive correlation. We find no significant correlation between BooookScore coherence scores and the writer coherence scores ($r$:.11, $p$:.18). The average BooookScore scores are almost the same across all three models (GPT-4: .95, Claude: .93, Llama: .93), which does not capture the significant difference in coherence that the writers judged in Llama in particular. We find significant correlation between FABLES and the writer faithfulness scores but the correlation is weak and less than just using GPT-4 ($r$: .29, $p$: .01). The average FABLES scores are also approximately the same across the three models (GPT-4: .99, Claude: .99, Llama: .99). Overall, these results support prior work \cite{chakrabarty2023art}, which shows that \textbf{LLMs are not yet reliable evaluators for skilled writing tasks}.

For context, we also include simple automatic metrics and recent Faithfulness-specific finetuned models. The only metric which performs better than GPT-4 is BERTScore for coherence using the story as a reference. However, the correlation is weak, and \citet{goyal2022snac} has already shown that BERTScore does not penalize many coherence errors. We additionally include the Llama-2-7B computed perplexity scores in this correlation table to check if there is any relationship between the similarity of stories to LLM training data and attribute ratings, and we find none.



\section{Discussion}
Our results show that, in the best summaries, models are capable of some interesting analysis of the themes present in unseen stories. When models are asked to summarize stories 
that may have analysis available online, it is not clear if the models are simply regurgitating analysis from the training data. Therefore, it is important to examine their capabilities on original and challenging content. Additionally, while we expected it might be challenging for LLMs to identify salient information within long stories, we find that long-context models demonstrate as good understanding of longer stories as shorter ones.

The models also have notable weaknesses. We demonstrate that even the best models are making significant errors across all summary attributes on about half of summaries. They are often too vague or missing important context, and they struggle with challenging subtext, providing unsupported analysis and misinterpreting character feelings/reactions and actions. This will be an important area for future study as it demonstrates language models still struggle with aspects of theory of mind\footnote{Theory of mind is the "ability to understand and keep track of the mental states of others" \cite{xu2024opentom}.}, one of the most challenging being understanding an unreliable narrator. Additionally, our analysis of LLM-based evaluation shows that they should not replace the expert human judgments. 


Working with writers was an efficient and mutually enjoyable 
method of ensuring we were not evaluating on any training data, and we were getting an informed evaluation of characteristics like faithfulness and analysis. Writers were able to complete summary ratings in a matter of minutes, whereas someone unfamiliar with the story would have taken much longer. 
For example, it took us, the paper authors, about 60-90 minutes to read and review the summaries for just one story, whereas it took the writers abot 5-10 minutes to do the same. Writers also left positive feedback like, "\textit{I'm glad to have read this [summary]... It shows some [weaknesses] of my story... some minor characters are more flat than I want.}"\\\\
\textbf{Limitations.} We provide a first look at how LLMs summarize unseen short stories within a reasonable cost and timeframe, but the work is limited in its number of stories, ratings, and prompting techniques. A follow-up study could expand these areas, but it is challenging given that an individual author has a limited number of high-quality stories that are complete but unpublished, there are a limited number of writers willing to participate in a study involving LLMs and corporate APIs, and compensation is expensive. As a result of these challenges, our numbers of stories and writers are similar to other studies which have used experienced creative writers as shown in Section \ref{sec:relatedwork}.\\\\
\textbf{Ethics.}
We follow protocol approved by our IRB for this study. In line with the TACL code of ethics, we protect writers' work and identities by saving data on secure servers attached only to anonymous IDs, approving any published excerpts from their work with them, and requesting our prompting data not be used for future model training. We also support the code of ethics by involving practitioners from the field influenced by our study in our work. One of the authors, Melanie Subbiah, has an equity interest in OpenAI.

\section{Conclusion}
We work with writers to provide unpublished short stories and evaluate the quality of LLM-generated summaries of these stories. We present a holistic evaluation at the span, summary, and story level of summary quality grounded in narrative theory that is based on data LLMs did not train on. We identify that LLMs can demonstrate understanding of long narrative and thematic analysis, but they struggle with specificity and reliable interpretation of subtext and narrative voice. Our methodology sets an important example of how we can collaborate with domain experts to reach beyond the paradigm of evaluating LLMs with LLMs on data they may have been trained on.

\section*{Acknowledgments}
We would like to express our gratitude to the short story writers for sharing their work and contributing annotations. Additionally, we would like to thank our reviewers and Action Editor for their thoughtful feedback.

\bibliography{tacl2021}
\bibliographystyle{acl_natbib}


\onecolumn

\appendix

\section{Information Sheet for Informed Consent}
\label{sec:consent}
\small
\textbf{Purpose of research study}: We are collecting judgments from writers of how well automatic summarization models can summarize the writers’ own short stories to study the level of understanding these models demonstrate of this type of writing.

\textbf{Benefits}: This study may not benefit you directly, although it may be interesting to see an AI model’s takeaways from your story. The study could, however, benefit society through identifying strengths/limitations of AI models when processing nuanced long text documents.

\textbf{Data protection}: We understand it is very important to protect your unpublished short stories. We will store your stories on password-protected Columbia University computers for the duration of this study and then delete them after completion of the study. We may like to include short excerpts (<1 paragraph) from your short stories and evaluation responses in an eventual publication as examples, but we will not do so without getting your permission at that time. Other than this, we will not publish your stories in any way. We do have to run your stories through the OpenAI, Anthropic, and HuggingFace company APIs to access some of the AI models. All three companies
state that they will not train on any data. The data may be retained at the company for 30-90 days to ensure consistency in their service but this data is not used or released by the company.

\textbf{Risks}: Since, we have to run your unpublished short story through the OpenAI, Anthropic, and HuggingFace company APIs, there is a small risk involved with trusting these companies are doing what they say they’re doing and preventing data leaks. Anthropic is additionally HIPPAA certified for safe data handling, and one of the authors, Melanie Subbiah, has run her own unpublished short stories through these APIs.

\textbf{Voluntary Participation}: You may stop participating at any time without penalty by emailing Kathleen McKeown (kathy@cs.columbia.edu). At that time, we will delete any short stories you may have shared with us. You will only be compensated for stories for which you complete the full evaluation process. We may end your participation if you are not following the instructions.

\textbf{Compensation}: You will be compensated \$50 for each short story/evaluation you complete in the form of a prepaid VISA card mailed to you.

\textbf{Confidentiality}: No identifying information will be published about you. We record whether you have published a short story before and/or completed a writing program to give a general sense of our study participants. Your mailing address will be kept for the duration of the study so that we can mail your payment card to you, and then promptly deleted. All of your work will be submitted in relation to an anonymous ID so we will not be able to connect your answers to you. The Columbia University Human Research Protection Office may obtain access to the data collected for this study.

\textbf{COI Disclosure}: Please be aware that Melanie Subbiah (an investigator on this study) has a financial interest in OpenAI, a company which makes some of the automatic summarization models which will be evaluated in this study.

\textbf{Questions/concerns}: You may email questions to the principal investigator, Kathleen McKeown: kathy@cs.columbia.edu. If you feel you have been treated unfairly you may contact the Columbia University Institutional Review Board and reference Protocol AAAU8875. Please respond to this email by copying the below consent paragraph to indicate that you understand the information in this consent form. You have not waived any legal rights you otherwise would have as a participant in a research study. If you have any questions about your rights or responsibilities as a research participant, please contact the Columbia University HRPO office at: Phone \{phone-number\}; Email \{email\}. We recommend either printing this information sheet or taking note of this contact information before beginning the study in case of questions.\\\\
\normalsize
“I have read the above purpose of the study, and understand my role in participating in the research. I volunteer to take part in this research. I have had a chance to ask questions. If I have questions later, about the research, I can ask the investigator listed above. I understand that I may refuse to participate or withdraw from participation at any time. The investigator may withdraw me at his/her professional discretion. I certify that I am 18 years of age or older and freely give my consent to participate in this study.”

\section{Model Prompts and Costs}
\label{sec:prompts}
\subsection{GPT-4}
\textbf{Cost} (experiments, summary generation, and evaluation): \$58\\\\
\textbf{Summary Generation}:\\
\textbf{System prompt}\\
You are an expert summary-writer. Summarize the provided passage in several paragraphs using only information from the passage provided.\\
\textbf{User prompt}\\ 
Title: \{title\}\textbackslash n\textbackslash nStory:\textbackslash n\{story\}\textbackslash n\textbackslash nWrite a coherent, chronological, and detailed but brief summary for this passage in several paragraphs (about 400 words). Briefly introduce key entities like characters or settings when they are mentioned in the summary, and include some analysis of the story.\textbackslash nSummary:\\\\
\textbf{Summary Evaluation}:\\
\textbf{System prompt}\\
You are a skilled writer and editor. Evaluate the quality of a summary of a short story by answering the questions provided. Select from the four options indicated and choose whichever fits best. Be careful to evaluate the summary in relation to the short story provided.\\
\textbf{User prompts} (questions are asked one by one and model answers are kept in the messages)
\begin{itemize}
\item Short Story: \textbackslash n\{story\}\textbackslash n\textbackslash nSummary:\textbackslash n\{summary\}
\item Does the summary cover the important plot points of the story?\textbackslash n1) No - critical details are left out that are necessary to understand the story\textbackslash n2) Not really - it would be hard to appreciate the story from the details provided\textbackslash n3) Mostly - covers the main points but small things missing\textbackslash n4) Yes - the important details of the narrative are covered
\item Does the summary misrepresent details from the story or make things up?\textbackslash n1) Yes - the summary includes incorrect details\textbackslash n2) Somewhat - the summary misrepresents details\textbackslash n3) Not really - mostly accurate but some details aren’t clear\textbackslash n4) No - everything is correct in relation to the story
\item Is the summary coherent, fluent and readable?\textbackslash n1) No - contains grammar errors or non sequiturs\textbackslash n2) Not really - confusing to follow but fluent\textbackslash n3) Mostly - a bit clunky but coherent and fluent\textbackslash n4) Yes - easy to read and understand
\item Does the summary provide any correct analysis of some of the main takeaways or themes from the story?\textbackslash n1) No - there is no analysis in the summary\textbackslash n\textbackslash n2) Not really - there is some analysis but it’s not correct\textbackslash n3) Somewhat - there is some correct analysis but it’s not very thoughtful\textbackslash n4) Yes - the summary touches on some of the themes/feelings/interpretation that you hoped to communicate as the writer"
\end{itemize}

\subsection{Claude}
\textbf{Cost} (on Claude limited free plan): \$0\\
\texttt{from anthropic import HUMAN\_PROMPT, AI\_PROMPT}\\
\textbf{Summary Generation}:\\
\textbf{Prompt}\\\{HUMAN\_PROMPT\} \{gpt4\_user\_prompt\}\{AI\_PROMPT\}\\\\
\textbf{Summary Evaluation}:\\
\textbf{Prompt} (questions asked one at a time in this format)\\\{HUMAN\_PROMPT\} \{gpt4\_system\_prompt\}\textbackslash n\textbackslash n\{gpt4\_storysummary\_user\_prompt\}\textbackslash n\textbackslash n\\\{gpt4\_question\_user\_prompt\}\textbackslash n\textbackslash nYou must place your score within <score><\textbackslash score>tags.\textbackslash n\textbackslash n\\{\{AI\_PROMPT\}\\\\

\subsection{Llama}
\textbf{Cost} (on HuggingChat): \$0\\
\textbf{Summary Generation}:\\
\textbf{System prompt}\\
 You are an expert summary-writer. Summarize the provided story in several paragraphs using only information from the story provided.\\
\textbf{Story-level (if the story is short enough) prompt}:\\
Write a coherent, chronological, and detailed summary for this story in several paragraphs (about 400 words). Briefly introduce key entities like characters or settings when they are mentioned in the summary, and include some analysis of the story.\\
\textbf{Chunk-level prompt}:\\Write a coherent, chronological, and detailed summary for this part of a story in several paragraphs (about 400 words). Briefly introduce key entities like characters or settings when they are mentioned in the summary, and include some analysis of the story.\\
\textbf{Concatenated-summaries prompt}:\\Compress these summaries of several parts of a story into a couple paragraphs (about 400 words) that are a coherent, chronological, and detailed summary of the whole story. Briefly introduce key entities like characters or settings when they are mentioned in the final summary, and include some analysis of the story.

\section{Short Story Uploader and Evaluation Interfaces}
\label{sec:interface}
See Figure \ref{fig:interface}. \\
\textbf{Cost}: \$1,250 to compensate writers for their work

\begin{figure}
    \centering
    \includegraphics[width=\textwidth]{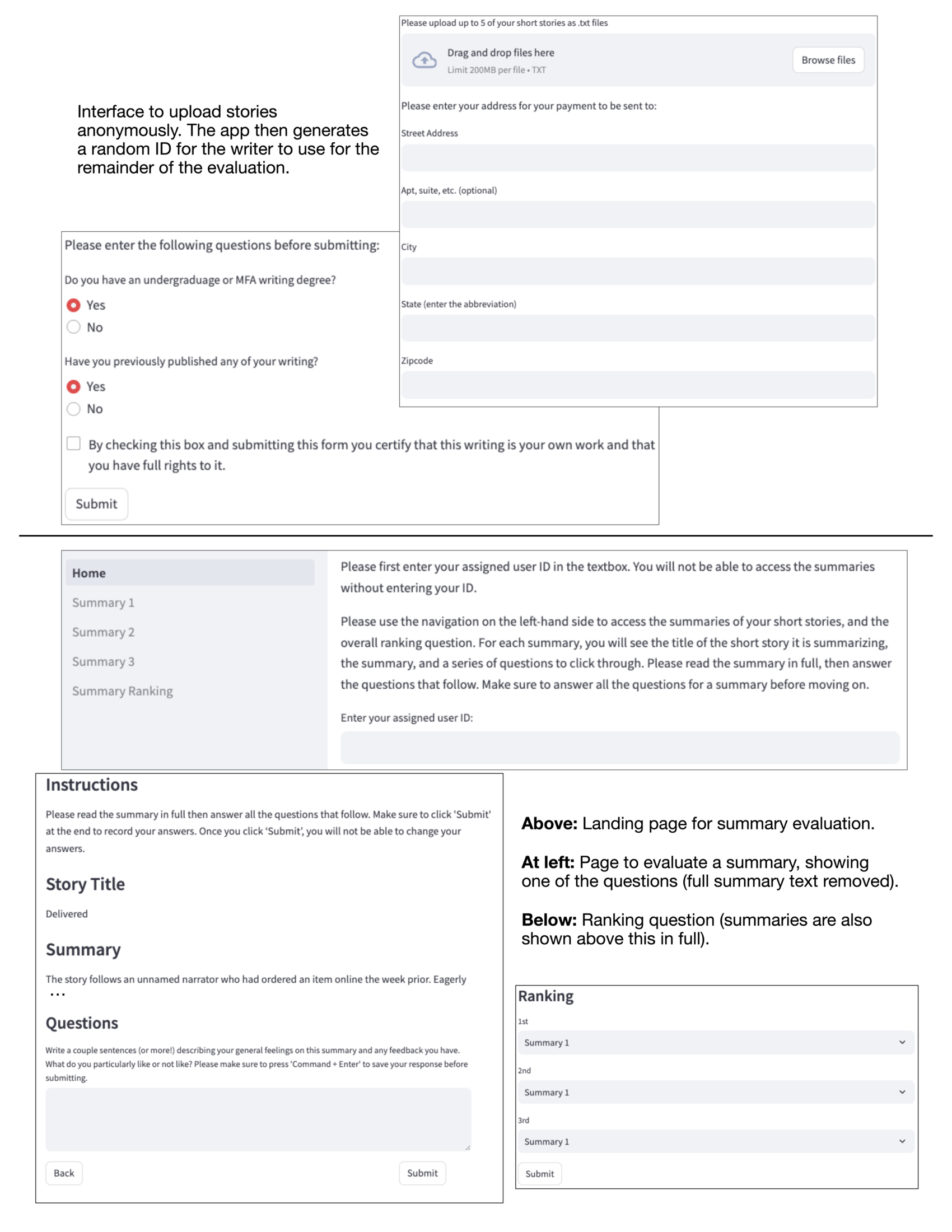}
    \caption{Screenshots of the \href{https://streamlit.io}{Streamlit} apps writers use to anonymously submit stories and evaluate summaries.}
    \label{fig:interface}
\end{figure}

\section{Fine-Grained Error Annotation Interface}
\label{sec:doccano}
See Figure \ref{fig:doccano}.\\
\textbf{Cost}: \$1,150 to compensate writers for their work
\begin{figure}
    \centering
    \includegraphics[width=\textwidth]{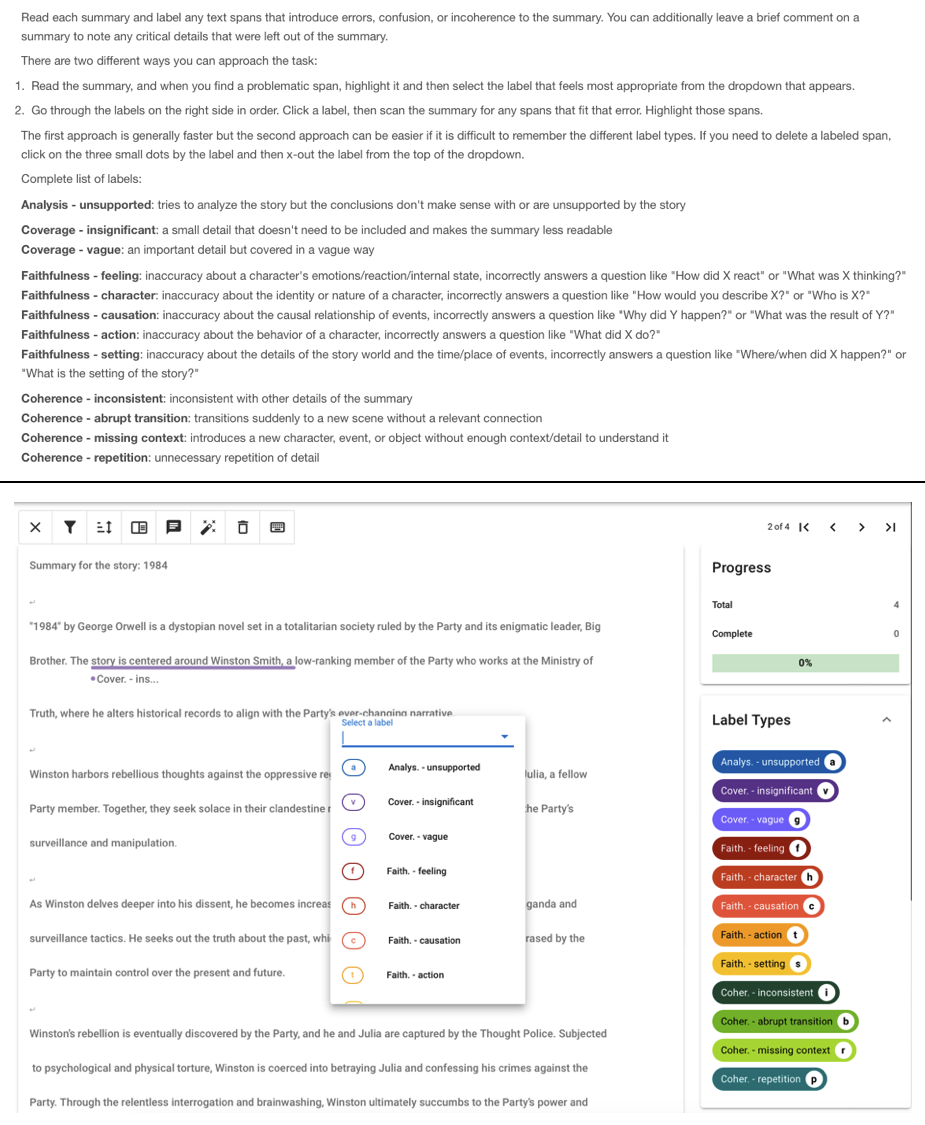}
    \caption{Top panel shows the instructions for the fine-grained error annotation task. Bottom panel shows the interface which was created using Doccano \cite{daudert2020web}. A user can highlight a span of text and then select a label from the dropdown to assign to it.}
    \label{fig:doccano}
\end{figure}

\end{document}